\documentclass[acmsmall]{acmart}

\AtBeginDocument{%
  }

\setcopyright{acmlicensed}
\copyrightyear{2025}
\acmYear{2025}
\acmDOI{XXXXXXX.XXXXXXX}

\acmJournal{JACM}
\acmVolume{37}
\acmNumber{4}
\acmArticle{111}
\acmMonth{8}

\usepackage{booktabs}    
\usepackage{makecell}    
\usepackage{multirow}
\usepackage{threeparttable}

\usepackage{amssymb}
\usepackage{pifont}
\usepackage{graphicx}
\usepackage{array}
\usepackage{rotating}
\usepackage{caption}
\usepackage{geometry}
\usepackage{subfigure}
\usepackage{changepage}
\usepackage{array}
\usepackage{cleveref}
\usepackage{utfsym}
\newcommand{\checkbox}{\usym{2610}}
\newcommand{\checkedbox}{\usym{2611}}
\usepackage{float}

\begin{document}

\title{A Survey on Efficiency Optimization Techniques for DNN-based Video Analytics: Process Systems, Algorithms, and Applications}

\author{Shanjiang Tang}
\authornote{Shanjiang Tang is the corresponding author.}
\email{tashj@tju.edu.cn}
\affiliation{%
  \institution{College of Intelligence and Computing, Tianjin University}
  \city{Tianjin}
  \postcode{300072}
  \country{China}
}

\author{Rui Huang}
\affiliation{%
  \institution{College of Intelligence and Computing, Tianjin University}
  \city{Tianjin}
  \postcode{300072}
  \country{China}
}

\author{Hsinyu Luo}
\email{iamxinyuluo@tju.edu.cn}
\affiliation{%
  \institution{College of Intelligence and Computing, Tianjin University}
  \city{Tianjin}
  \postcode{300072}
  \country{China}
}

\author{Chunjiang Wang}
\email{chunjiang_wang@mail.ustc.edu.cn}
\affiliation{%
  \institution{School of Biomedical Engineering, USTC}
  \city{Suzhou}
  \state{Jiangsu}
  \country{China}
}

\author{Ce Yu}
\email{yuce@tju.edu.cn}
\affiliation{%
  \institution{College of Intelligence and Computing, Tianjin University}
  \city{Tianjin}
  \postcode{300072}
  \country{China}
}

\author{Yusen Li}
\email{liyusen@nbjl.nankai.edu.cn}
\affiliation{%
  \institution{School of Computing, Nankai University}
  \city{Tianjin}
  \postcode{300071}
  \country{China}
}

\author{Hao Fu}
\email{fuhao@nscc.tj.cn}
\affiliation{%
  \institution{National Supercomputing Center of Tianjin}
  \city{Tianjin}
  \country{China}
}

\author{Chao Sun}
\email{sch@tju.edu.cn}
\affiliation{
  \institution{College of Intelligence and Computing, Tianjin University}
  \city{Tianjin}
  \postcode{300072}
  \country{China}
}

\author{Jian Xiao}
\email{xiaojian@tju.edu.cn}
\affiliation{
  \institution{College of Intelligence and Computing, Tianjin University}
  \city{Tianjin}
  \postcode{300072}
  \country{China}
}

\renewcommand{\shortauthors}{Trovato et al.}

\begin{abstract}
  The explosive growth of video data in recent years has brought higher demands for video analytics, where accuracy and efficiency remain the two primary concerns. Deep neural networks (DNNs) have been widely adopted to ensure accuracy; however, improving their efficiency in video analytics remains an open challenge. Different from existing surveys that make summaries of DNN-based video mainly from the accuracy optimization aspect, in this survey, we aim to provide a thorough review of optimization techniques focusing on the improvement of the efficiency of DNNs in video analytics. We organize existing methods in a bottom-up manner, covering multiple perspectives such as hardware support, data processing, operational deployment, etc. Finally, based on the optimization framework and existing works, we analyze and discuss the problems and challenges in the performance optimization of DNN-based video analytics.
\end{abstract}



\received{12 June 2025}
\received[revised]{--}
\received[accepted]{--}

\maketitle

\section{Introduction}
Video has been viewed as a dominant medium for information dissemination. In daily life, the widespread popularity of short videos further accelerates the development of video analytics, applied in numerous domains that require real-time video processing such as autonomous vehicles~\cite{21video,17reverse}, drones~\cite{18Urban,18bandwidth,21drones} or surveillance video analysis~\cite{20Lightweight, 19Intelligent, 20DLApproach}. According to statistics, mobile video traffic accounts for 79\% of total mobile traffic between 2017 and 2022~\cite{19Cisco}, which underscores the growing importance of video analysis. As such, video analysis tasks must be effectively addressed with two critical evaluation criteria: accuracy and efficiency. Therefore, the central challenge in video analytics lies in extracting needed information accurately and efficiently.\par

Efficiency has received growing attention in DNN-based video analytics due to the massive volume of video data and the demand for low-latency responses in real-time applications. While DNN architectures such as AlexNet~\cite{12AlexNet}, ResNet~\cite{16ResNet}, and VGGNet~\cite{14VGG} are widely used as backbone networks that effectively ensure accuracy, the emergence of AI, 5G, and IoT technologies, along with the proliferation of real-time interactive video applications~\cite{21real,18bandwidth,21drones}, has raised user expectations for responsiveness, imposing stricter efficiency requirements. As a result, novel optimization techniques should be continually explored to enhance the execution efficiency of video analytics applications.\par


Given the importance of efficiency metrics, significant research efforts have been devoted to optimizing the efficiency of DNN-based video analytics. One line of work focuses on improving storage and computational efficiency by fundamentally optimizing the underlying storage systems and computational resources. This includes techniques such as dynamic resource scheduling and optimized storage coding, which aim to maximize model execution efficiency under resource constraints. Another line of research targets operational efficiency from a system deployment perspective, particularly in real-world applications such as smart IoT. These studies aim to reduce video streaming load and latency by leveraging cloud-based solutions, including edge compression models and edge–cloud co-computing frameworks. New iterations of hardware and software technologies as well as new or higher application requirements from users bring new challenges and opportunities for efficiency optimization of DNN-based video analytics, which indicates the necessary for comprehensively summarizing and discussing the efficiency optimization techniques for DNN-based video analytics.\par

\begin{figure}[htb]
    \centering
    \includegraphics[width=13.8cm]{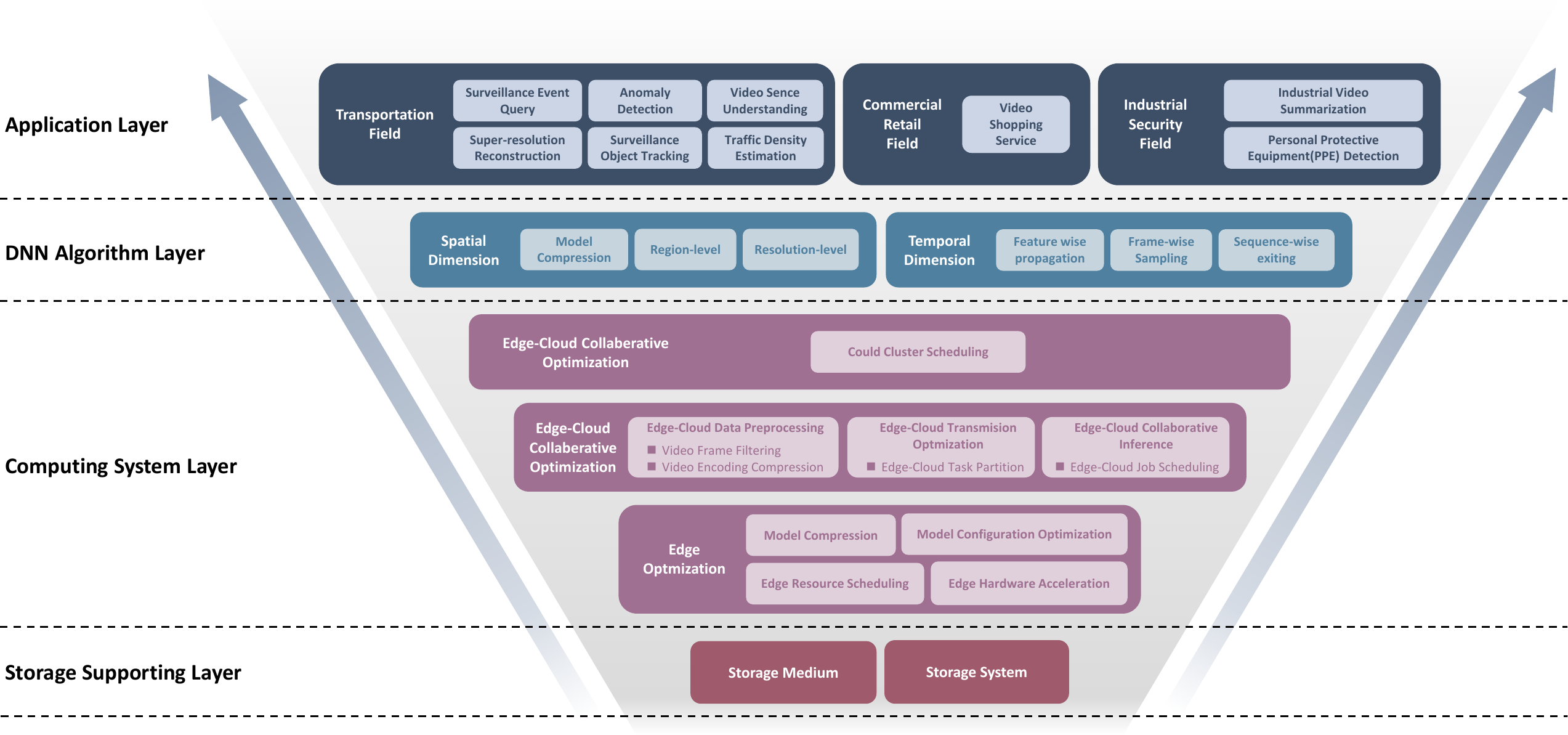}
    \caption{Overview of the efficiency optimization work for DNN-based video analytics from the bottom to the up. We divide it into four layers, i.e., storage supporting layer, computing system layer, DNN algorithm layer and application layer.}
    \label{fig:DetailLayers}
\end{figure}

Recently, several reviews of DNN-based video analytics have been conducted.  Some of them~\cite{22human, 20review, 18video, 22violence, 21topic, 08video} compare and summarize the contribution of multiple deep learning model architectures to analytic accuracy in specific visual analytics tasks (e.g., video prediction, video scene analysis), specific video types (e.g., sports videos), or specific video application domains, respectively while others ~\cite{19edge, 17survey, 20overview, 22deep} discuss the issue of efficiency optimization in video analytics. However, these reviews just summarize efficiency optimization techniques from a certain stage of video analytics (e.g., model deployment stage). Different from the above-mentioned reviews, this paper is devoted to summarizing the efficiency optimization techniques of DNN-based video analytics systematically with different layers considered (see Fig.~\ref{fig:DetailLayers}).\par

The aim of this paper includes two parts. First, comprehensively review the latest efficiency optimization techniques for DNN-based video analytics to establish a foundational cognitive system. A bottom-up approach is employed to review relevant works from hardware, deployment, algorithm, and application optimization, across four layers in the optimization system: storage supporting, computing system, DNN algorithm, and application layer. Second, we discuss future trends and challenges in optimizing the efficiency of DNN-based video analytics, drawing insights from in-depth studies.\par

The remainder of the paper is organized as follows (see Fig.\ref{fig:DetailLayers}). \Cref{sec:StorageSupport} describes the storage optimization for DNN-based video analytics from the perspective of storage media and storage system, respectively. \Cref{sec:SysArchitecture} covers various system performance optimization techniques from different perspectives, including edge-side and edge-cloud deployment modes. \Cref{sec:DNNAlgorithm} summarizes feasible methods for optimizing the inference performance of DNN-based algorithms, considering both temporal and spatial dimensions. \Cref{sec:Applicaion} introduces video application optimization approaches around application scenarios. \Cref{sec:Challenges} examines the opportunities and challenges of efficiency optimization for DNN-based video analytics. Finally, \Cref{sec:Conclusion} concludes the paper.


\section{Storage Supporting Layer}\label{sec:StorageSupport}

Coupled with advances in DNN-related technologies, the growing demand for real-time video processing has intensified the need for storage efficiency, as data read/write latency and storage capacity limitations are decisive factors. In this section, we cover storage efficiency optimization techniques for DNN-based video analytics from the hardware(storage medium) and software(storage systems) aspects.

\subsection{Storage Medium}

The design and selection of storage medium directly impact storage efficiency. Volatile memory is susceptible to energy leakage, whereas non-volatile memory exhibits lower leakage power but is burdened by write latency and energy consumption. To overcome these limitations, hybrid storage solutions have emerged, capitalizing on the strengths of both memory types. AMBER scheme~\cite{13amber}, based on NVM and SRAM, an integrated energy management solution for video applications, uses a block-matching algorithm combined with an on-chip frame buffer to reduce the read/write latency of the external memory. Furthermore, EnHyV~\cite{14energy}, which combines a small SRAM array with STT-RAM, reduces bit-toggling during write operations and supports parallelized video processing compared to AMBER. Volatile STT-RAM~\cite{16exploring} decreases video write latency and energy by reducing the retention time of the magnetic tunnel junction (MTJ).


\subsection{Storage Systems}

The process of video storage includes transcoding, storage, decoding, and computation. A key component in this process is the video database management system (VDBMS), which orchestrates the storage and retrieval of video data on storage media. Current research on enhancing storage efficiency with VDBMS mainly focuses on compressing storage space and improving data access performance, outlined in Table \ref{storagetable}.
\begin{itemize}
  \item \emph{VisualCloud~\cite{17visualcloud}, LightDB~\cite{18lightdb}}: 
  Both LightDB and VisualCloud focus on optimizing access to virtual reality (VR) data by predicting the proximity of user's view in time and decoding specifically for the predicted direction, yet their applicability to other video types is limited.
  \item \emph{Vignette~\cite{19perceptual}}: As a cloud-based perceptual storage manager, Vignette integrates the human perceptual system into the cloud storage infrastructure. By merging perceptual insights with video transcoding pipelines, it creates perception-based video compression, efficiently reducing storage space at large scales. But it requires additional training for perception information extraction modules.
  \item \emph{VStore~\cite{19vstore}}: 
  VStore accelerates data queries by adaptively configuring video formats(e.g. fidelity, encoding) that are adjusted automatically to facilitate multi-version video caching and reduce frequent transcoding/decoding operations, but constrained by the storage demands of diverse formats along with prerequisites for knowledge.
  \item \emph{VSS~\cite{21vss}}: 
   The design of VSS storage systems considers how high-level video operations affect low-level storage details, including optimizing disk data layout, caching frequently accessed subregion data, and eliminating video overlap areas via feature vectors. This enhances access efficiency and saves storage space. Nonetheless, deploying these advanced optimizations necessitates profound professional knowledge and technical expertise.
   \item \emph{TASM~\cite{21tasm}}: 
   Apart from other optimization strategies, TASM focuses on optimizing spatial frame subset access as a tile-based layout video data storage manager. It dynamically divides data into fine granularity based on query volume and video content, and adaptively adjusts tiled layouts for rapid spatial random access; however, this optimization requires balancing the re-tiling time and the access speed enhancement achieved. 
\end{itemize}

\begin{table}[htb]
  \centering
  \caption{Comparison of Different Video Storage Systems}
  \Description{This table compares six video storage systems based on data type, access method, storage architecture, and optimization focus.}
  \label{storagetable}
  \renewcommand\arraystretch{1.5}
  \begin{tabular}{ccccc}
    \toprule
    \midrule
    \textbf{\makecell[c]{System}} &
    \textbf{\makecell[c]{Data\\Type}} &
    \textbf{\makecell[c]{Access\\Method}} &
    \textbf{\makecell[c]{Storage\\Architecture}} &
    \textbf{\makecell[c]{Optimization\\Focus}} \\
    \midrule
    VisualCloud~\cite{17visualcloud} & VR & SA & Distributed & DA \\
    LightDB~\cite{18lightdb} & VR & SA & Distributed & DA, SS \\
    Vignette~\cite{19perceptual} & SV, VR & SA & Distributed & SS \\
    VStore~\cite{19vstore} & SV & SA & Single-node & DA \\
    VSS~\cite{21vss} & SV & SA & Single-node & DA, SS \\
    TASM~\cite{21tasm} & SV & RA & Single-node & DA \\
    \midrule
    \bottomrule
  \end{tabular}

  \vspace{0.5em}
  \hspace*{0.12\linewidth} 
  \begin{minipage}{0.9\linewidth}
    \footnotesize
    VR = virtual reality video data; SV = standard video data. \\
    SA = sequential access; RA = random access. \\
    SS = compress storage space; DA = improve data access.
  \end{minipage}
  
\end{table}

\section{Computing System Layer}\label{sec:SysArchitecture}

\begin{figure}[htb]
    \centering    
    \includegraphics[width=7.5cm]{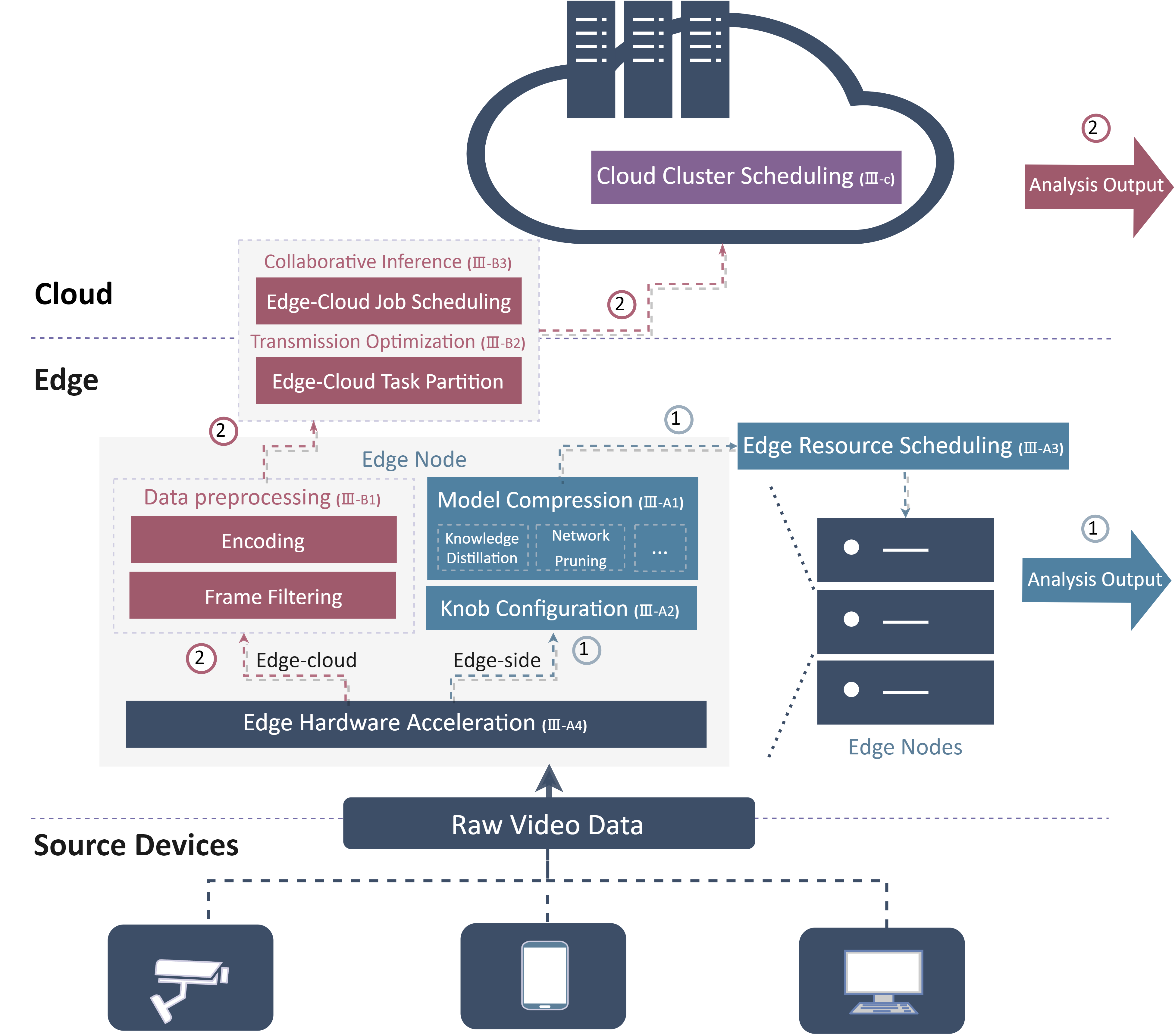}
    \caption{Overview of computing system optimization. \textcircled{1}: Edge optimization path; \textcircled{2}: Edge-Cloud optimization path. The content within parentheses indicates the positional relationship between each subsection header and the main text.}
    \label{fig:computing system}
\end{figure}

Given the localized nature of optimizing storage efficiency, it is crucial to incorporate real-world operational paradigms such as edge computing to explore opportunities for system efficiency improvements further. Modules in video analysis systems (e.g., encoding, decoding, detection) leverage the flexibility and convenience of modern technology, allowing for decentralized deployment across various locations (source, edge, or cloud). Yet, this location choice involves a trade-off between data locality and computing capacity: edge resources are limited but close, while cloud resources are abundant but distant, affecting overall analysis efficiency. To cope with these challenges, researchers have proposed various system performance optimization techniques from different perspectives, as shown in Table \ref{technologytable}. Furthermore, in this section, accounting for the trade-off between data locality and computing capacity, we will discuss research results related to edge, edge-cloud, and cloud data processing levels (see Fig.\ref{fig:computing system}).

\subsection{Edge Optimization}

The advancement of edge computing capabilities has driven the shift of video analysis tasks from cloud to edge nodes, accelerating processing and enhancing user privacy protection. For example, in scenarios~\cite{16edge} such as locating a lost child, transmitting large-scale camera data to the cloud introduces privacy risks and delay. In contrast, edge devices in the search area directly address local camera data from cloud requests, quickly generating the final result at the near end, then returning only the outcome to the cloud. Nevertheless, this process may be constrained by the limited resources on the edge device, impacting overall processing performance.\par

\begin{table}[htbp]

    \centering
    \caption{Efficiency Optimization Techniques for Edge or Edge-Cloud Systems}
    \label{technologytable}
    \renewcommand\arraystretch{1.0}
    \renewcommand{\cellgape}{\Gape[3.3pt]}
    \resizebox{\linewidth}{!}{
        \begin{tabular}{
            >{\centering\arraybackslash}m{2.2cm} 
            >{\centering\arraybackslash}m{2.8cm} 
            >{\centering\arraybackslash}m{3.8cm} 
            >{\centering\arraybackslash}m{2.2cm} 
            >{\centering\arraybackslash}m{5.3cm} 
            >{\centering\arraybackslash}m{5.1cm}
        }
            \toprule
            \midrule
            \textbf{Mode} & 
            \textbf{Optimization Type} & 
            \textbf{Technique} & 
            \textbf{Year} & 
            {\centering\textbf{Methodologies}} & 
            \textbf{Optimization Objectives} \\
            \midrule

            \multirow{30}{*}{\shortstack{Edge \\Optimization}}

            & \multirow{12}{*}{\makecell{Model \\Compression}}
            & \makecell[c]{Parameter Quantization \\ \cite{18inference,chang2018energy,21dyQuantization}} & \makecell{2018, 2021}
            & \makecell[l]{
                \textbullet\ Compress bit width\\
                \textbullet\ Adaptive quantize precision}
            & \multirow{12}{*}{
                \begin{tabular}[t]{@{}ll@{}}
                    \checkedbox\ Accuracy   & \checkbox\ Utilization \\
                    \checkedbox\ Latency    & \checkedbox\ Cost \\
                    \checkbox\ Throughput   & \checkbox\ Bandwidth
                \end{tabular}
            } \\
            \cline{3-5}
            
            && \makecell{Network Pruning \\ \cite{20effective,19pruning}} & 2019-2022
            & \makecell[l]{
                \textbullet\ Structured pruning\\
                \textbullet\ Cut redundant connections} & \\
            \cline{3-5}
            && \makecell[c]{Tensor Decomposition \\ \cite{21hybrid,20deepeye,21compressing}} & 2020, 2021
            & \makecell[l]{
                \textbullet\ Reconstruct weight matrix\\
                \textbullet\ Low-rank subspace} & \\
            \cline{3-5}
            && \makecell[c]{Knowledge Distillation \\ \cite{19online,222+,23mobilevos}} & 2019-2023
            & \makecell[l]{
                \textbullet\ Teacher-student mode\\
                \textbullet\ Train small model}& \\
            \cline{3-5}
            && \makecell[c]{Lightweight Structure \\Design \cite{20real,23scgnet}} & 2020, 2023
            & \makecell[l]{ 
                \textbullet\ Lightweight design\\
                \textbullet\ Deep separable convolutio} & \\
            \cline{2-6}

            & \multirow{11}{*}{\makecell{Model \\Configuration \\Optimization}}
            & ApproxDet \cite{20approxdet} & 2020
            & \makecell[l]{ 
                \textbullet\ Approximation computing\\
                \textbullet\ Content and contention-aware}
            & \multirow{11}{*}{
                \begin{tabular}[t]{@{}ll@{}}
                    \checkedbox\ Accuracy & \checkedbox\ Utilization \\
                    \checkedbox\ Latency  & \checkedbox\ Cost(energy) \\
                    \checkbox\ Throughput & \checkbox\ Bandwidth
                \end{tabular}
            } \\
            \cline{3-5}
            && FastAdapt \cite{21benchmarking} & 2021
            & \makecell[l]{
                \textbullet\ Heterogeneous resource analysis} & \\
            \cline{3-5}
            && SMARTADAPT \cite{22smartadapt} & 2022
            & \makecell[l]{
                \textbullet\ Fine-grained tuning\\
                \textbullet\ Video content-aware selection} & \\
            \cline{3-5}
            && Edge-Serverless Co-optimization \cite{wang2025edge} & 2025
            & \makecell[l]{
                \textbullet\ Serverless resources\\
                \textbullet\ Video knobs optimization} & \\
            \cline{3-5}
            && EdgeEye \cite{22edgeeye} & 2022
            & \makecell[l]{
                \textbullet\ Combinatorial optimization\\
                \textbullet\ Simplify configuration search} & \\
            \cline{2-6}

            & \multirow{4}{*}{\makecell[c]{Edge Resource \\Scheduling}}
            & \makecell[c]{Multi-Nodes Load \\Balancing \cite{20distream,21elf,18cooperative}} & 2018-2021
            & \makecell[l]{
                \textbullet\ Task-specific partitioning\\
                \textbullet\ Incentive allocation}
            & \multirow{4}{*}{
                \begin{tabular}[t]{@{}ll@{}}
                    \checkedbox\ Accuracy & \checkedbox\ Utilization \\
                    \checkedbox\ Latency  & \checkedbox\ Cost \\
                    \checkedbox\ Throughput & \checkedbox\ Bandwidth
                \end{tabular}
            } \\
            \cline{3-5}
            && \makecell[c]{Single-Node Multi-Tenant \\Scheduling \cite{18mainstream,18nestdnn}} & 2018
            & \makecell[l]{
                \textbullet\ Computation sharing\\
                \textbullet\ Resource-accuracy tradeoffs} & \\
            \cline{2-6}

            & \makecell[c]{Edge Hardware \\Acceleration}
            & \makecell[c]{Edge AI Hardware \\ \cite{NCS,GoogleCoral}} & 2018
            & \makecell[l]{
                \textbullet\ Acceleration hardware\\
                \textbullet\ Matrix calculation}
            & \multirow{1.5}{*}{
                \begin{tabular}[t]{@{}ll@{}}
                    \checkedbox\ Latency  & \checkedbox\ Cost
                \end{tabular} 
            } \\
            \hline

            \multirow{36}{*}{\makecell{Edge-Cloud\\Collaborative\\Optimization}}

            & \multirow{7}{*}{\makecell{Video\\Frame\\Filtering}}
            & \makecell[c]{FilterForward \\ \cite{18inference,chang2018energy,21dyQuantization}} 
            & \makecell{2019}
            & \makecell[l]{
                \textbullet\ Wide-area edge filtering \\
                \textbullet\ Frame-level classification}
            & \multirow{7}{*}{
                \begin{tabular}[t]{@{}ll@{}}
                    \checkedbox\ Accuracy   & \checkbox\ Utilization \\
                    \checkedbox\ Latency    & \checkbox\ Cost \\
                    \checkedbox\ Throughput   & \checkedbox\ Bandwidth
                \end{tabular}
            } \\
            \cline{3-5}

            & & \makecell[c]{Reducto \cite{20reducto}} 
            & \makecell[c]{2020} 
            & \makecell[l]{
                \textbullet\ Time-varying correlation \\
                \textbullet\ Frame differencing}
            & \\ 
            \cline{3-5}
            
            & & \makecell[c]{SmartFilter \cite{22smartfilter}} 
            & \makecell{2022} 
            & \makecell[l]{
              \textbullet\ Server-driven filtering \\
              \textbullet\ Lightweight binary classifier
              }
            & \\ 
            \cline{2-6}
            
            & \multirow{15}{*}{\makecell[c]{Video\\Encoding\\Compression}}
            & \makecell[c]{VideoChef \cite{18videochef}} 
            & \makecell{2018} 
            & \makecell[l]{
              \textbullet\ Canary input \\
              \textbullet\ Estimate quality degradation
              }
            & \multirow{15}{*}{
                \begin{tabular}[t]{@{}ll@{}}
                    \checkedbox\ Accuracy   & \checkbox\ Utilization \\
                    \checkedbox\ Latency    & \checkbox\ Cost \\
                    \checkbox\ Throughput   & \checkedbox\ Bandwidth
                \end{tabular}
            } \\
            \cline{3-5}
            
            & & \makecell[c]{ACCMPEG \cite{22accmpeg}} 
            & \makecell{2022} 
            & \makecell[l]{
              \textbullet\ Macroblock-level encoding \\
              \textbullet\ Server-driven
            }
            & \\ 
            \cline{3-5}
            
            & & \makecell[c]{AWStream \cite{18awstream}} 
            & \makecell{2018} 
            & \makecell[l]{
              \textbullet\ Accuracy-bandwidth tradeoffs \\
              \textbullet\ Adaptive coding parameters
            }
            & \\ 
            \cline{3-5}
            
            & & \makecell[c]{Cloudseg \cite{19bridging}} 
            & \makecell{2020} 
            & \makecell[l]{
              \textbullet\ Low quality encoding \\
              \textbullet\ Super-resolution procedure
            }
            & \\ 
            \cline{3-5}
            
            & & \makecell[c]{DDS \cite{20server}} 
            & \makecell{2020} 
            & \makecell[l]{
              \textbullet\ Key region encoding \\
              \textbullet\ Secondary inference
            } 
            & \\ 
            \cline{3-5}
            
            & & \makecell[c]{VPaaS \cite{21serverless}} 
            & \makecell{2021} 
            & \makecell[l]{
              \textbullet\ Fog-assisted encoding
            } 
            & \\ 
            \cline{3-5}

            & & \makecell[c]{CMC~\cite{song2024cmc}} 
            & \makecell{2024} 
            & \makecell[l]{
              \textbullet\ identify redundant tokens
            } 
            & \\ 
            \cline{2-6}

            & \multirow{8}{*}{\makecell[c]{Edge-Cloud\\Task\\Partition}}
            & \makecell[c]{mVideo \cite{19mvideo}} 
            & \makecell[c]{2019} 
            & \makecell[l]{
              \textbullet\ Video task phase partition \\
              \textbullet\ Edge and cloud collaboration
            } 
            & \multirow{8}{*}{
                \begin{tabular}[t]{@{}ll@{}}
                    \checkedbox\ Accuracy   & \checkbox\ Utilization \\
                    \checkedbox\ Latency    & \checkedbox\ Cost(energy) \\
                    \checkedbox\ Throughput   & \checkedbox\ Bandwidth
                \end{tabular}
            } \\
            \cline{3-5}
            
            & & \makecell[c]{Partial DI \cite{19embedded}} 
            & \makecell{2019} 
            & \makecell[l]{
              \textbullet\ Layer-by-layer partition \\
              \textbullet\ Task surge-oriented
            } 
            & \\ 
            \cline{3-5}
            
            & & \makecell[c]{Optimal DNN \\Partition  \cite{17neurosurgeon,19dynamic,19couper,21allocating}} 
            & \makecell{2017-2021} 
            & \makecell[l]{
              \textbullet\ Incorporate dynamic factors \\
              \textbullet\ DAG topology model \\
              \textbullet\ Adaptive scaling
            } 
            & \\ 
            \cline{2-6}
            
            & \makecell[c]{Edge-Cloud\\Job\\Scheduling}
            & \makecell[c]{MCDNN \cite{16mcdnn} \\ Deepdecision \cite{18deepdecision}} 
            & \makecell{2016, 2018} 
            & \makecell[l]{
              \textbullet\ Optimal offloading strategy \\
              \textbullet\ Combinatorial optimization \\
              \textbullet\ Balance accuracy and utilization
            }
            & \multirow{2}{*}{
                \begin{tabular}[t]{@{}ll@{}}
                    \checkedbox\ Accuracy   & \checkedbox\ Utilization \\
                    \checkedbox\ Latency    & \checkedbox\ Cost(energy) \\
                \end{tabular}
            } \\
            \hline

            \multirow{1}{*}{\raisebox{0.7ex}{\makecell[c]{Cloud\\Optimization}}}
                & \raisebox{-1.8ex}{\makecell[c]{Cloud Cluster\\Scheduling}}
                & \raisebox{-0.9ex}{\makecell[c]{Nexus \cite{19nexus} \\ DQN based \cite{18learningPara}}}
                & \raisebox{-1.8ex}{\makecell[c]{2018, 2019}}
                & \makecell[l]{
                  \textbullet\ Batch-adaptive \\
                  \textbullet\ Hierarchical scheduling \\
                  \textbullet\ Reinforcement learning
                }
                & \multirow{1}{*}{
                    \raisebox{-0.6ex}{
                      \begin{tabular}[t]{@{}ll@{}}
                          \checkedbox\ Utilization  & \checkedbox\ Latency \\
                          \checkedbox\ Cost(energy) & \checkedbox\ Throughput \\
                      \end{tabular}
                    }
                } \\


            \midrule
            \bottomrule
        \end{tabular}
        
    } 
    
    \vspace{0.5em}
      \hspace*{-0.1\linewidth} 
      \begin{minipage}{0.9\linewidth}
        \footnotesize
        "Cost" without parentheses refers to computing or storage costs.\\
        "Bandwidth" refers to the network capacity for data.
      \end{minipage}
\end{table}

To guarantee anticipated performance and align with edge device constraints, following path \textcircled{1} in Fig.\ref{fig:computing system}, our focus will be on two key optimization aspects for edge-side analytics: storage efficiency (e.g., model compression) and computational efficiency (e.g., model configuration optimization, edge resource scheduling).

\subsubsection{Model Compression} \label{sec:ModelCompression}

Compared to clouds, edge-side analytics account for longer latency and larger resource requirements that should be further considered. The scale of DNN-based models directly impacts runtime storage efficiency. Relevant studies identify that model compression is an effective method to address this bottleneck. By adjusting the model's parameters or structure, it is possible to significantly reduce model storage requirements and runtime overhead. \par

\textbf{Parameter Quantization.} Parameters of DNN are commonly represented by high-precision data type (e.g., 32-bit floating-point) that result in a large amount of storage overhead. Parameter quantization is applied to tackle this issue via compromising parameter bit width. For instance, Preußer et al.~\cite{18inference} explore parameter quantization, which generates binary neural networks~\cite{16binarized} (binarizing model weights and activation values) to satisfy resource-constrained heterogeneous embedded platforms. Inspired by the efficiency boost from quantization, VideoIQ~\cite{21dyQuantization}, a video instance-aware dynamic quantization framework, adaptively selects the optimal quantization precision based on frame segment content, strategically allocating computation by the temporal redundancy of video data. Meanwhile, Chang et al.~\cite{chang2018energy} devise a dataflow optimization method that quantizes the model parameters from 32-bit floating-point to 13-bit fixed-point representations, effectively reducing energy consumption while preserving high video quality. They further devise a dataflow optimization method that transforms CNN into the deconvolutional neural networks (DCNN) to tackle super-resolution (SR) input.\par

\textbf{Network Pruning.} In video analytics, network pruning focuses on diminishing redundant connections and parameters to reduce network complexity, which is inspired by the pruning idea of removing similar convolutional results among consecutive frames to accelerate inference. This method is typically classified into two aspects: unstructured pruning and structured pruning. Given that unstructured pruning requires additional algorithms and hardware support to reduce storage overhead~\cite{19deep}, here, we mainly discuss structured pruning techniques. Tsai et al.~\cite{20effective} proposed a filter pruning method that devotes to breaking the sparse matrix connection pattern in traditional pruning. Dai et al.~\cite{21compressing} focus on performance degradation in deep structures caused by uniform pruning and introduce an adaptive channel pruning technique based on naive Bayesian inference. Distinctly, Wang et al.~\cite{19pruning} introduce filter pruning for 3D convolution networks (involve the temporal information of sequential frames) to effectivly select candidate filter subsets.\par

\textbf{Tensor Decomposition.} Tensor decomposition compresses deep neural networks by restructuring weight matrices—decomposing low-rank matrices into high-dimensional tensors. This approach is particularly effective for the training process of video spatiotemporal models, where the highly non-convex optimization landscape often leads to excessive redundant parameters, causing slow convergence and overfitting. To address it, Dai et al.~\cite{21hybrid} employ tensor decomposition to factorize the weight matrix and reduce the parameter size. Additionally, other studies~\cite{20deepeye, 21compressing} apply tensor decomposition during inference. DEEPEYE~\cite{20deepeye} integrates LSTM-based tensor decomposition with tensorized temporal features to replace raw video inputs and achieve lightweight terminal deployment. At the same time, aiming for enhanced compression~\cite{21compressing}, pruning coupled with Tucker decomposition is employed to improve the compression ratio of the pruning model.\par

\textbf{Knowledge Distillation.} Knowledge distillation relies on the "teacher-student model" concept to alleviate resource strains, where a smaller "student model" approximates a larger "teacher model" through knowledge transfer. In terms of learning model capabilities, various solutions have been proposed from different supervisory perspectives. Mullapudil et al.~\cite{19online} adopt an online distillation method combined with self-supervised learning. It intermittently uses the teacher model's video object segmentation results as targets to guide the student model, reducing the cost of offline pre-training. And Miles et al.~\cite{23mobilevos} introduce a boundary-aware semi-supervised distillation technique, benefiting from pixel-level contrastive learning and pre-trained teacher distillation. Besides, Vu et al.~\cite{222+} propose an unsupervised distillation paradigm considering intermediate layer features as knowledge and further extracting knowledge from 2D and 3D teacher models. \par
  
\textbf{Lightweight Structure Design.} Lightweight structure design incorporates specialized structures into the network architecture to simplify model complexity. For instance, MobileNet's deep separable convolution efficiently reduces model complexity to a resource-acceptable state. Younis et al.~\cite{20real} combine MobileNet lightweight network with SSD (Single-Shot MultiBox Detector) to detect complex video scenes on embedded devices. For the limited receptive field problem of MobileNet, which struggles with capturing large-scale frame information, Zhang et al.~\cite{23scgnet} introduce the SCG lightweight network. It utilizes cascaded and shifted group convolutions to capture multi-scale information.

\subsubsection{Model Configuration Optimization} \label{sec:sysEdge_Md}
Certain video system frameworks employ approximation algorithms that reduce computational load to adaptively infer within a limited timeframe while switching suboptimal model configurations (e.g., input resolution, detector interval, the number of proposals). Since offline trained models often fail to meet changing performance requirements, ApproxDet\cite{20approxdet} takes dynamic video content, resource availability, and user latency-accuracy requirements into account to build a multi-branch model with approximate configurations. Likewise, FastAdapt~\cite{21benchmarking}, a variant of ApproxDet, explores the implications of heterogeneous resource contention and power consumption on configuration selection through comprehensive benchmark tests. Furthermore, SMARTADAPT~\cite{22smartadapt} builds upon ApproxDet and FastAdapt, discussing optimal selection within these branch sets and proposing fine-grained execution branch switching based on content-aware information.\par 

Recent studies extend configuration tuning to serverless-edge environments. Wang et al.~\cite{wang2025edge} jointly optimized video knobs (frame rate, DNN model size) and serverless resources (memory allocation) through a Markov approximation algorithm. By dynamically selecting configurations per time slot and triggering keyframes via frame-difference analysis, their approach reduced computation costs by 89.37\% while guaranteeing target accuracy. This demonstrates the scalability of configuration optimization in hybrid computing paradigms. \par

Traditionally, the achievement of optimal configurations required periodic analysis. However, even small adjustments to each configuration can lead to myriad possible permutations. In response, EdgeEye~\cite{22edgeeye} employs data-driven techniques to transform the edge configuration selection problem into a combinatorial optimization problem (COP),  simplifying the search scope.\par

\subsubsection{Edge Resource Scheduling}
Within edge environments, scheduling primarily concentrates on efficiently allocating limited edge resources and minimizing negative co-located tasks' contention while maintaining each task's performance.
\newline
\indent \textbf{Multi-Node Load Balancing.}
  The gap between computing demands and available computing resources at the edge can be bridged by balancing the workload among edge nodes to ensure efficient node cooperation and minimize resource fragmentation. Some studies exhibit high flexibility in load partitioning. For instance, Distream~\cite{20distream} partitions load based on the classification logic of task object attributes, while Elf~\cite{21elf} specifically considers video frame data as partition objects and proposes a content-aware frame partitioning algorithm to allocate load across multiple computing nodes.  
   Moreover, studies like~\cite{18cooperative}, discuss real-time task assignment mechanisms for heterogeneous edge nodes, balance resource allocation among tasks, and incentivize node resources.\par 
  
  \textbf{Single-Node Multi-Tenant Scheduling.} In this scenario, multiple tasks share fixed edge resources concurrently. Maximizing concurrent performance, Mainstream~\cite{18mainstream} focuses on partial DNN computation sharing across multiple tasks that dynamically adjusts the level of backbone sharing based on the tradeoffs between task specialization and sharing. 
  NestDNN~\cite{18nestdnn} concentrates on the dynamic availability of runtime resources and automatically achieves the best resource-accuracy trade-off for each deep learning model. \par

\subsubsection{Edge Hardware Acceleration}
Abundant computing resources are the basis for improving system efficiency. Recently, emerging AI hardware provides specialized acceleration for edge video tasks. The Neural Compute Stick (NCS)~\cite{NCS} is powered by a Vision Processing Unit (VPU) for efficient inference at low power consumption. Meanwhile, several studies~\cite{19detailed, 20face} have demonstrated the significant advantages of utilizing NCS for video inference on mobile and embedded vision devices. So far, there have been some successful commercial products such as the Intel Neural Compute Stick 2 (NSC2) and the Google Coral USB accelerator~\cite{GoogleCoral} with an edge Tensor Processing Unit (TPU) core. However, large-scale edge deployment of AI accelerators involves a trade-off between implementation costs and efficiency gains. \par

\subsection{Edge-Cloud Collaborative Optimization} \label{sec:CollaborativePro}
Similar to edge-side analytics, cloud-involved analytics also adapts to unique scene characteristics like handling larger-scale data, meeting higher precision demands, and processing data in real time despite stricter bandwidth limitations.\par

Further considering the larger-scale searching for lost children, centralized comparison of video data across search areas necessitates robust computing power in the far cloud. In this case, edge devices assist cloud search, transmitting key information to the cloud, which then further analyzes and generates wide-area search results. This process consistently relies on high-quality video data to fulfill tasks, often with assistance from the edge.\par

\begin{figure}[htb]
    \centering
    \includegraphics[width=7cm]{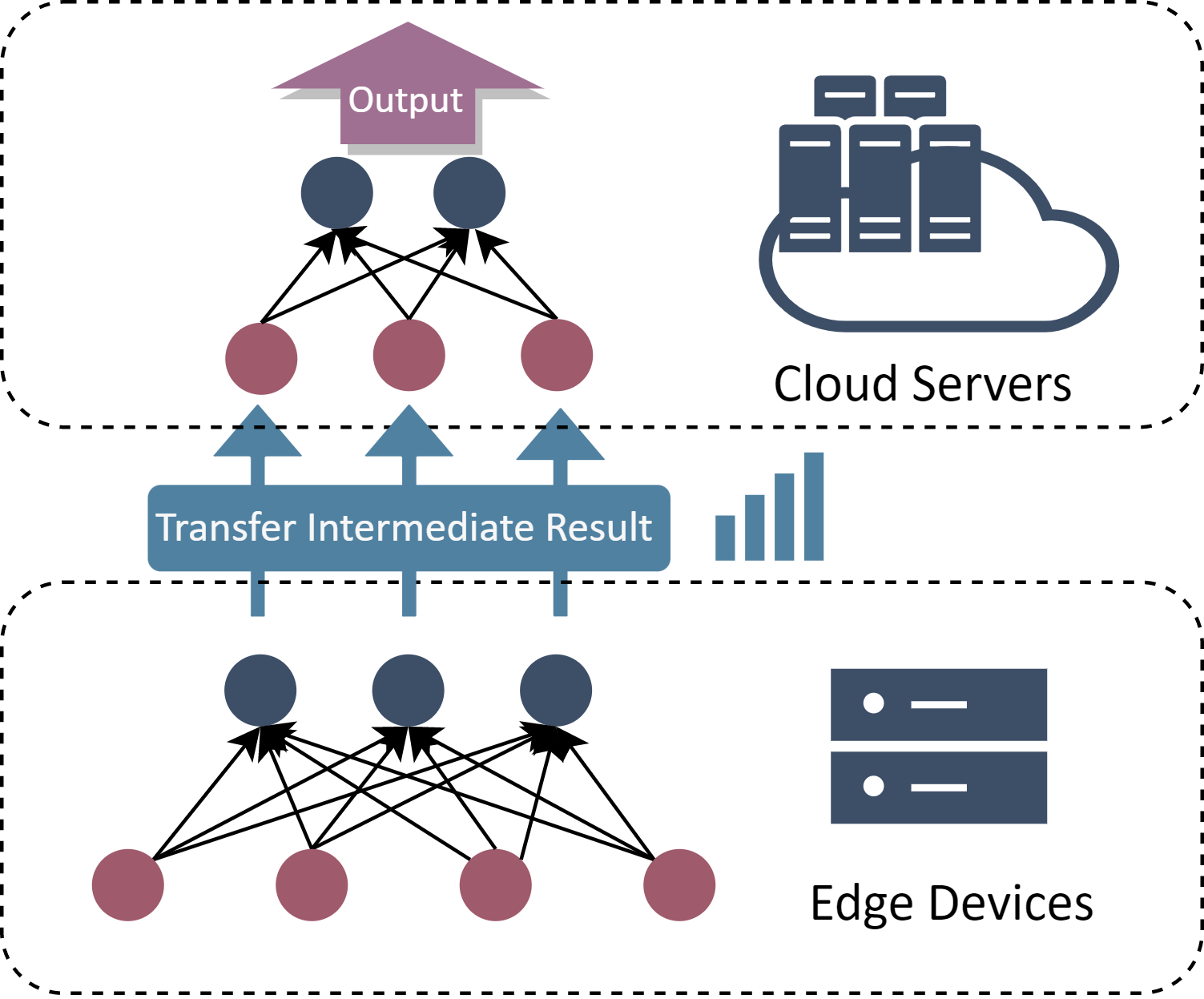}
    \caption{Edge-Cloud task partition workflow.}
    \label{fig:DNNpartition}
\end{figure}

\subsubsection{Edge-Cloud Data Preprocessing}
When processing video data in the cloud, increasing the number of frames analyzed by intensive models raises computation costs and degrades operational efficiency. To address this issue, applying pre-processing steps at the edge, such as edge filtering and video encoding, helps reduce data transmission and alleviate cloud workload.\par

\textbf{Video Frame Filtering.} 
 Considering the severe bandwidth limitations and heavy processing pressure of transmitting the entire video to the cloud in real-time applications, studies like FilterForward~\cite{19scaling} employs a DNN-based lightweight edge filter that focuses on filtering highly correlated frames, where more correlated frames contribute more to the result. Reducto~\cite{20reducto}, a source filtering system that fully utilizes the computational ability of source devices (e.g., cameras), employs a frame difference filter based on low-level features at the source. Differently, SmartFilter~\cite{22smartfilter}, an early adopter of server-driven filtering, dedicatedly improves accuracy loss caused by filtering based on server-side task feedback. It also tackles the poor interpretability of frame difference vectors and scalability limitations faced by Reducto.\par

\textbf{Video Encoding Compression.} Several studies have investigated the impact of encoding on performance, proposing diverse video frame codec channels based on various perceptual sources. VideoChef~\cite{18videochef}, from the perspective of video content perception, is inspired by guiding encoding configuration with small input. It estimates the degradation in quality resulting from complete inputs by the observed small input accuracy errors. Furthermore, AccMPEG~\cite{22accmpeg} focuses on smaller and finer encoding areas, implementing precise encoding for each (16x16) macroblock based on server-side DNN accuracy impact. DDS (DNN-Driven Streaming)~\cite{20server}, considers the uneven distribution problem of important pixels between frames and proposes the adaptive "secondary inference" by re-encoding the critical areas.
Bandwidth and Task-Aware Encoding~\cite{18neural} argues that encoding should be co-designed with an eventual perception objective. It proposes bandwidth-aware encoders (shallow DNNs). Meanwhile, AWStream~\cite{18awstream} researches the adaptation of codecs to varying network conditions. It explores the bandwidth consumption-analysis accuracy tradeoffs and efficiently compresses video by setting the parameters (resolution, QP, and frame rate) of the underlying codec. Cloudseg~\cite{19bridging} transmits a video at a lower quality and then enhances the video using super-resolution on the server side. Compared to AWStream, CloudSeg reduces bandwidth consumption by 6.8x. The VPaaS system~\cite{21serverless} joins widely deployed fog nodes to further optimize the encoding quality. In contrast to methods that tune codec parameters for transmission efficiency, recent work like CMC~\cite{song2024cmc} proposes a novel perspective: reusing the internal mechanisms of hardware video codecs to assist inference tasks directly. By leveraging the macroblock-level motion estimation engine in codecs, CMC identifies temporally and spatially redundant tokens in video Transformers, enabling dynamic pruning during inference. Though primarily designed for edge-side acceleration, this approach reveals a promising direction where codecs are co-optimized not just for bandwidth but also for DNN workload reduction—paving the way for deeper integration of video encoding and computation in edge-cloud collaborative pipelines.\par

\subsubsection{Edge-Cloud Transmission Optimization}
To reduce data transmission between the edge and the cloud, multiple studies have explored methods for edge-cloud task partitioning.\par

\textbf{Edge-Cloud Task Partition.} Greatly smaller than the raw input, intermediate data in DNN layer propels the way of edge-cloud task partition to reduce computation resources overhead. Typically, the workflow (see Fig.~\ref{fig:DNNpartition}) runs partitioned DNNs on edge devices, uploading intermediate results to the cloud for further analysis. Studies like mVideo~\cite{19mvideo} emphasize the collaboration between edge nodes and clouds that segment video task phases based on available resources at edge nodes and transmit intermediate results to the cloud. Dey et al.~\cite{19embedded} tackle blocking-prone scenarios caused by task surge through employing a layer-by-layer partitioning algorithm based on the deep inference graph, maximizing the number of processable layers to achieve partial depth inference while minimizing intermediate storage. \par 

Notably, beyond network load consideration, recent adoption of directed acyclic graph (DAG) structures in DNNs has complicated the selection of partitioning points, which prompts extensive search in this field. Xing et al.~\cite{21allocating} focus on dynamic DNN partitioning for parallel processing tasks that formalizes the problem to a selection of optimal DNN partitioning points, which takes the minimum latency as the objective function and considers the network congestion, server load, and other states. Neurosurgeon~\cite{17neurosurgeon} favors predicting the latency/energy cost of individual DNN layers based on their configuration and type. It likewise incorporates dynamic factors (e.g., network bandwidth, load) to select the optimal point. Dynamic Adaptive DNN Surgery (DADS) scheme~\cite{19dynamic} ameliorates Neurosurgeon's performance degradation for the DAG topology model and further adaptively divides DNN layers according to the load level. Building upon the aforementioned work, Couper~\cite{19couper} identifies the challenge of numerous potential splitting points caused by complex models and proposes the custom slicing method to rapidly partition DNNs into appropriately sized components.\par

\subsubsection{Edge-Cloud Collaborative Inference}Certain simple tasks can be inferred at the edge while complex tasks require the support of robust computing power in the cloud. Collaborative inference between edge devices and cloud servers enables the efficient sharing of computation, storage, and network resources by selecting the optimal computing location based on task requirements and characteristics. \par

\textbf{Edge-Cloud Job Scheduling.}
Considering the available resources at the edge and cloud, some systems employ heuristic methods to search for optimal scheduling strategies by maximizing (or minimizing) an objective function under constraints. For example, MCDNN~\cite{16mcdnn} explores effective scheduling methods for heterogeneous request streams, aiming to balance accuracy and efficiency within resource limitations by scheduling variant models and selecting execution locations (edge or cloud). In the DeepDecision framework~\cite{18deepdecision}, the offloading problem is formalized as a combinatorial optimization problem that determines the optimal offloading strategy based on the impact of metric variations on execution locations, with specific metrics including network bandwidth, latency, accuracy, and so on.

\subsection{Cloud Optimization} 
\textbf{Cloud Cluster Scheduling.} Ensuring optimal throughput for video tasks within single-task constraints is a major concern in large-scale cloud-based video analytics. Although cloud clusters provide sufficient computing power, the expected throughput improvement often fails to meet expectations due to improper task allocation. Batch-based scheduling distributes the startup load on computational resources, enhancing throughput performance. However, oversized batches can raise processing latency. Nexus Cluster Engine~\cite{19nexus}, from the perspective of optimizing batch processing, uses a batch-aware scheduler to choose the optimal inference batch size and automate the allocation of GPU resources to adjust latency boundaries for DNN video tasks. In addition, studies like~\cite{18learningPara} are devoted to solving the unbalanced resource utilization in hybrid cloud environments, caused by uncertain task characteristics and heterogeneous resources. It delves into hierarchical scheduling based on reinforcement learning, training a two-stage scheduling model for efficiently assigning tasks in a fine-grained manner. \par

\section{DNN Algorithm Layer}\label{sec:DNNAlgorithm}
\begin{figure}[htb]
    \centering
    \includegraphics[width=12.5cm]{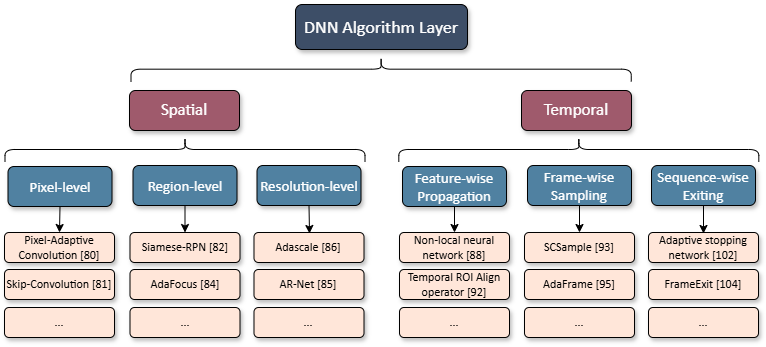}
    \caption{Improving DNN algorithm efficiency through spatial and temporal optimization.}
    \label{DNN tree}
\end{figure}

 The gradual improvement of DNN inference algorithms tailored for video analysis, coupled with optimization support at the computing system level, further releases the potential of efficiency optimization. Inspired by the spatio-temporal duality of video data itself, this section aims to summarize viable methods for optimizing the inference performance of DNN-based algorithms from temporal and spatial perspectives (see Fig.\ref{DNN tree}). Note that temporal and spatial optimization are orthogonal and can simultaneously exist.

\subsection{Spatial Dimension}
\begin{figure}[htb]
  \centering
  \begin{subfigure}[Pixel-level optimization]{
    \includegraphics[scale=0.075]{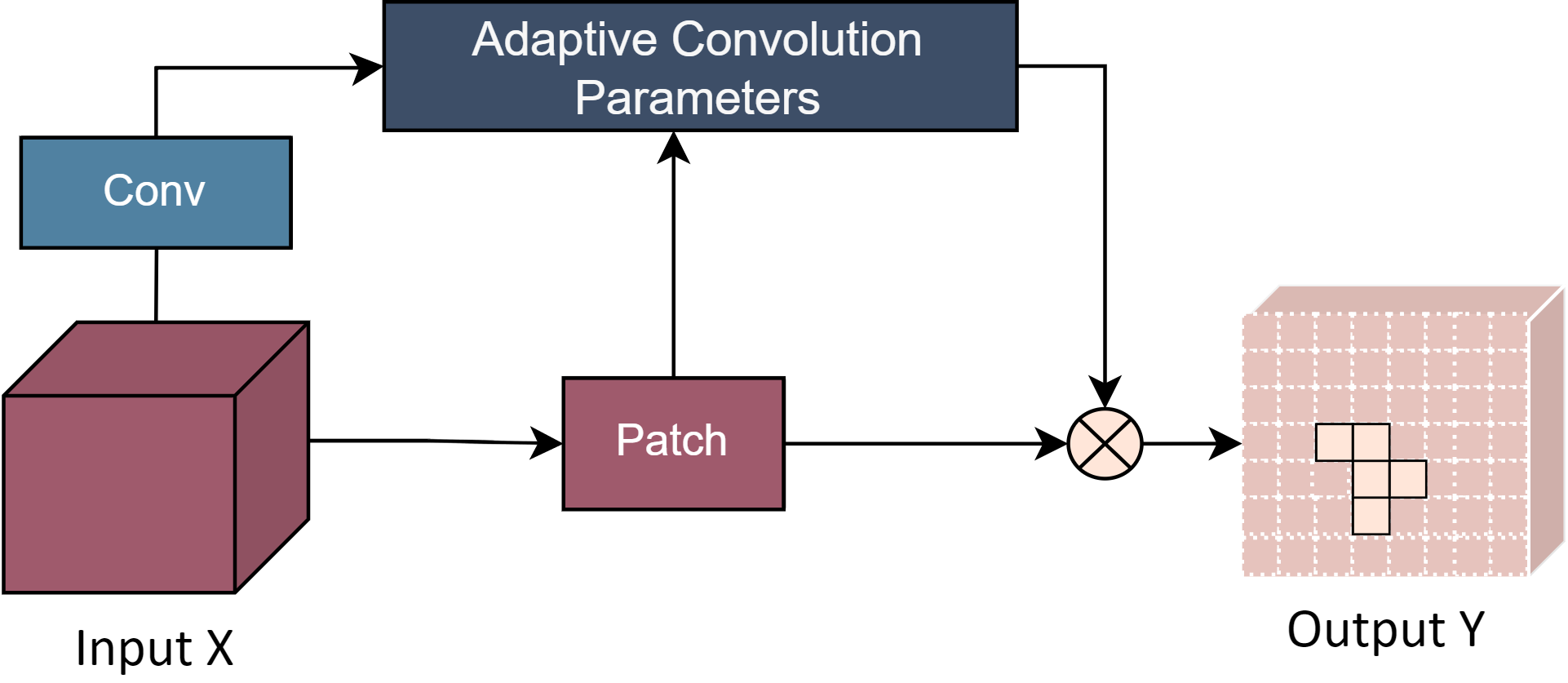}
    \label{pixel}}
 \end{subfigure}
 \hspace{10mm}
  \begin{subfigure}[Region-level optimization]{
    \includegraphics[scale=0.075]{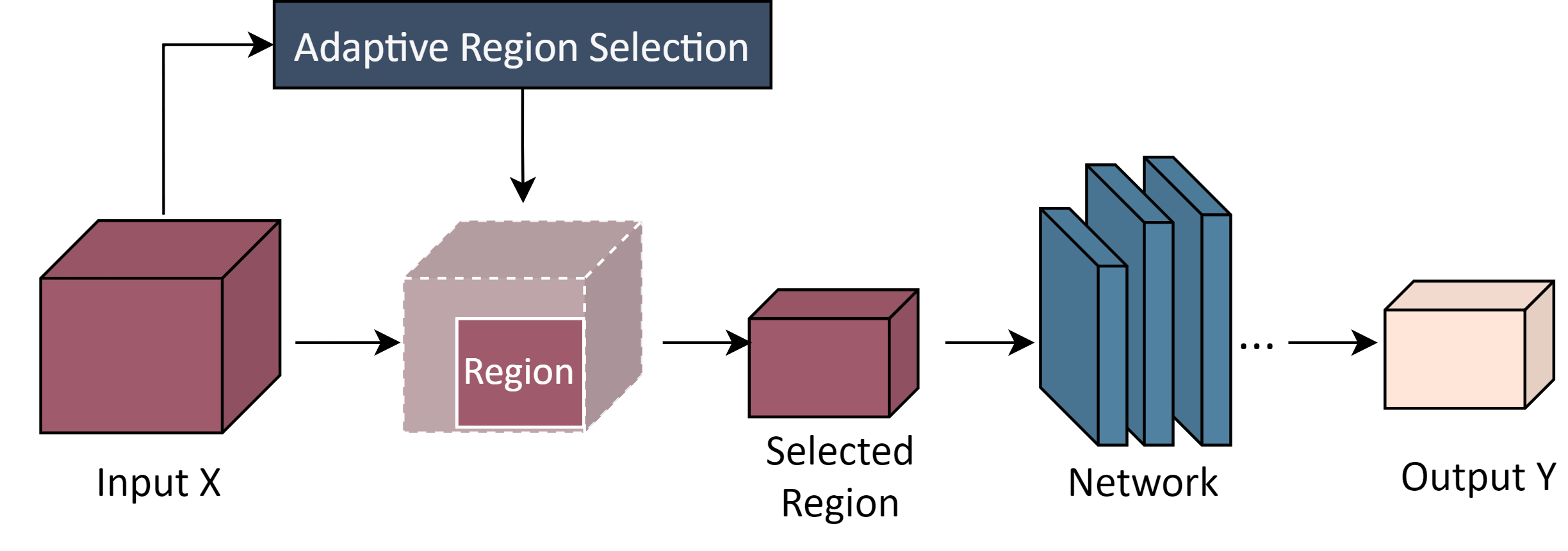}
    \label{region}}
 \end{subfigure}
 \hspace{10mm}
  \begin{subfigure}[Resolution-level optimization]{
    \includegraphics[scale=0.075]{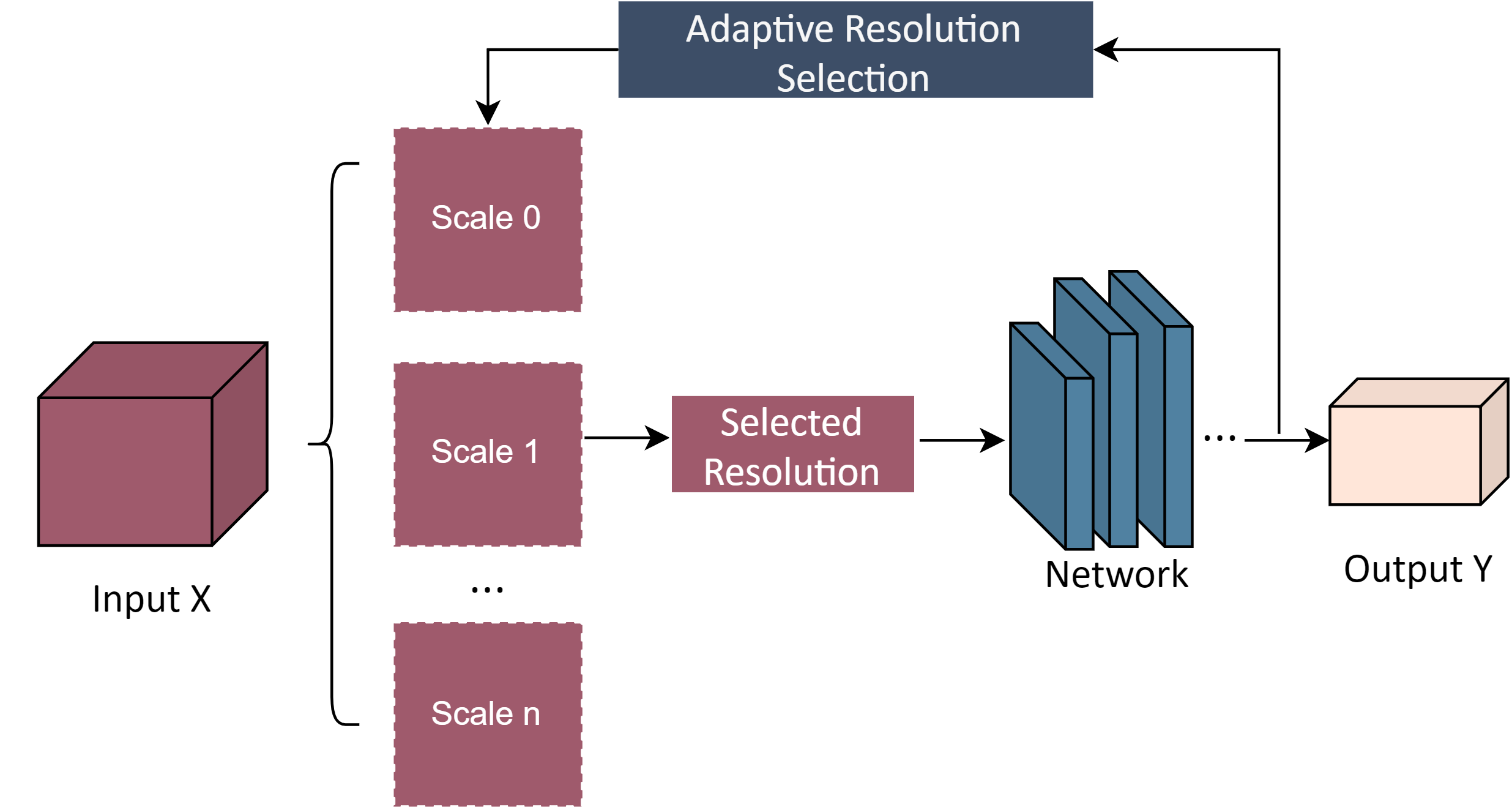}
     \label{resolution}}
  \end{subfigure}
  \caption{DNN-Based spatial dimension optimization focuses on: (a) Pixel-level: adaptive convolution parameters. (b) Region-level: adaptive region selection. (c) Resolution-level: adaptive resolution.}
  \label{spatial dimension}
 \end{figure}

In response to the unevenness of information distribution in video frames, spatial dimension optimization~\cite{19pixel,21skipconvolutions,18high,21patchnet,21adaptive,20ar,19adascale,19patchwork} employs techniques such as pixel cropping, region partitioning, and frame quality reduction to focus on crucial areas and diminish spatial redundancy. Based on frame granularity, we further categorize these into pixel-level (see Fig.\ref{pixel}), region-level (see Fig.\ref{region}), and resolution-level (see Fig.\ref{resolution}).


\subsubsection{Pixel-level Optimization} The human visual system processes a constant stream of visual data but can efficiently focus on essential elements, enabling effortless awareness of the environment. Following this idea, pixel-level optimization dynamically adjusts weight parameters to appropriately enhance the calculation of specific pixels or uses masking functions to improve the model's focus within the input space.\par

\textbf{Adaptive Convolution Parameters.}  
Pixel-Adaptive Convolutional Neural Networks (PACNN)~\cite{19pixel} emphasize fixed convolutions within CNNs and leverage Pixel-Adaptive Convolution (PAC) as its key innovation, which employs predefined or learned features to construct adaptive kernels. Skip-Convolution~\cite{21skipconvolutions} proposes that adaptive convolution should account for variations in inter-frame information and network activations. It restricts the computation to the significant variation parts through gating masks, while sparsifying the convolution computation for the non-significant parts.

\subsubsection{Region-level Optimization}Region-level optimization techniques (e.g., adaptive region selection) aim to efficiently reduce redundancy by centering computations on high-value regions while minimizing or disregarding low-value regions.\par

\textbf{Adaptive Region Selection.} Li et al.~\cite{18high} propose two proposal selection strategies: discarding region proposals generated by distant anchors, and reordering proposal scores using cosine window and scale change penalties. They develop the Siamese region proposal network (Siamese-RPN), consisting of a Siamese subnetwork for feature extraction and a region proposal subnetwork with classification regression. Inspired by similar low-level features across consecutive frames, Mao et al.~\cite{21patchnet} improve the efficiency of Siamese-RPN by transforming pixel-level feature learning to patch-level feature learning. Wang et al.~\cite{21adaptive} observe that the most informative region is usually a small image patch, which shifts smoothly across frames. Based on this, they develop an adaptive focus (AdaFocus) method based on reinforcement learning to select the most valuable regions of interest (ROI), which employs a lightweight CNN to aware coarse global information and locates relevant ROIs through a policy network. Furthermore, for latency-sensitive applications, Chai~\cite{19patchwork} only processes small regions per frame. He adopts a q-learning-based training strategy to intelligently select local regions and utilize memory units to retain the context of region loss, thus improving prediction efficiency based on previous frames and current local updates.\par

\subsubsection{Resolution-level Optimization}
Different from pixel or region-level optimization, coarse-grained resolution optimization reduces the amount of information processed by adaptively adjusting the resolution of video frames.\par

\textbf{Adaptive Resolution.} Adascale~\cite{19adascale} focuses on the similarities between consecutive frames in a video sequence and proposes an adaptive scaling technique that evaluates the performance of each pre-defined scale (resolution) by combining regression and classification losses. It leverages a regression classifier to select the best-performing scale. And AR-Net~\cite{20ar}, an adaptive resolution network, consists of a policy network that determines the resolution and different backbone networks corresponding to different resolutions.\par

\subsection{Temporal Dimension}
\begin{figure}[htb]
  \centering
  \begin{subfigure}[Feature-wise propagation]{
    \includegraphics[scale=0.057]{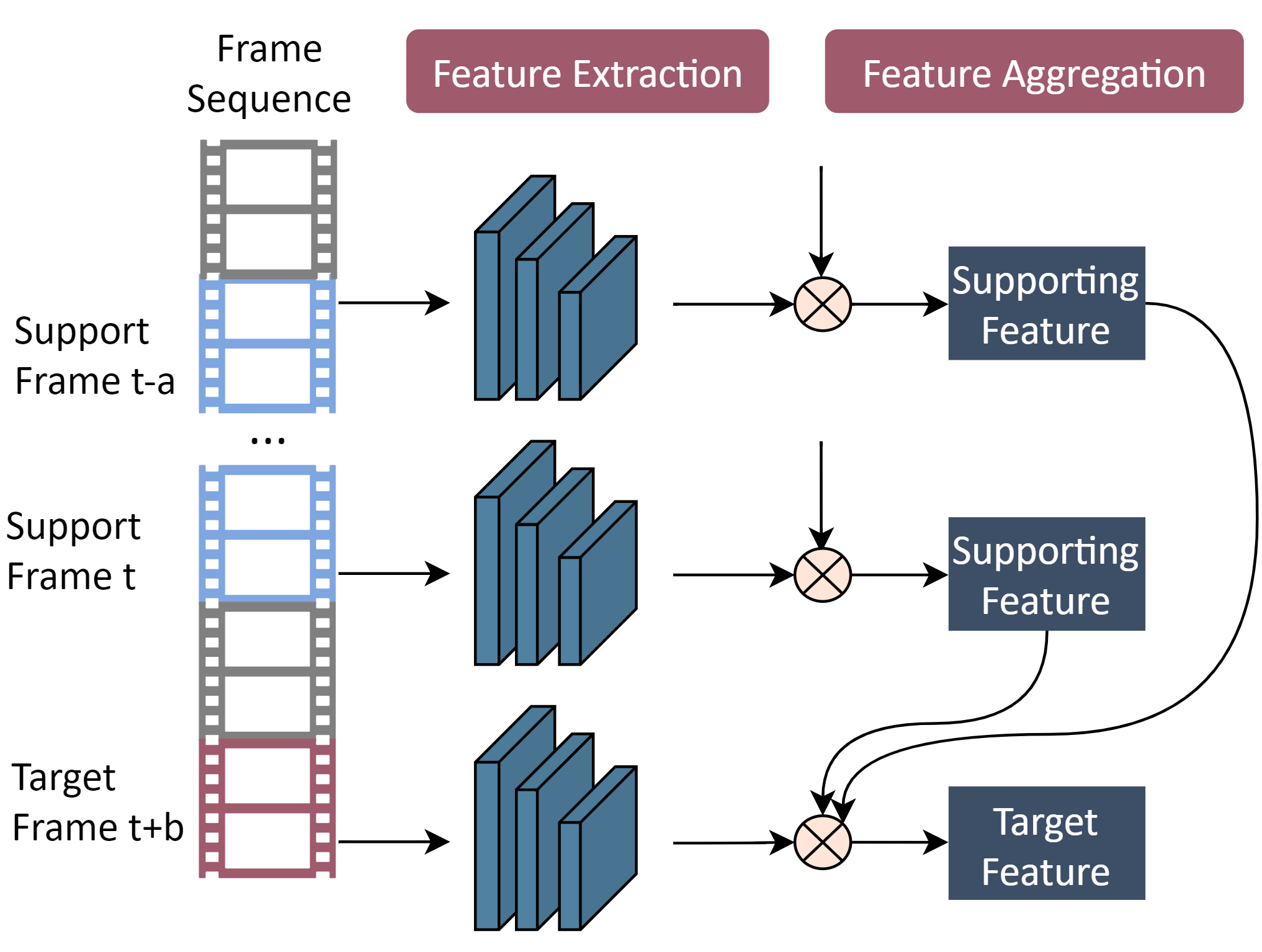}
    \label{longRange}}
 \end{subfigure}
  \begin{subfigure}[Frame-wise sampling]{
    \includegraphics[scale=0.058]{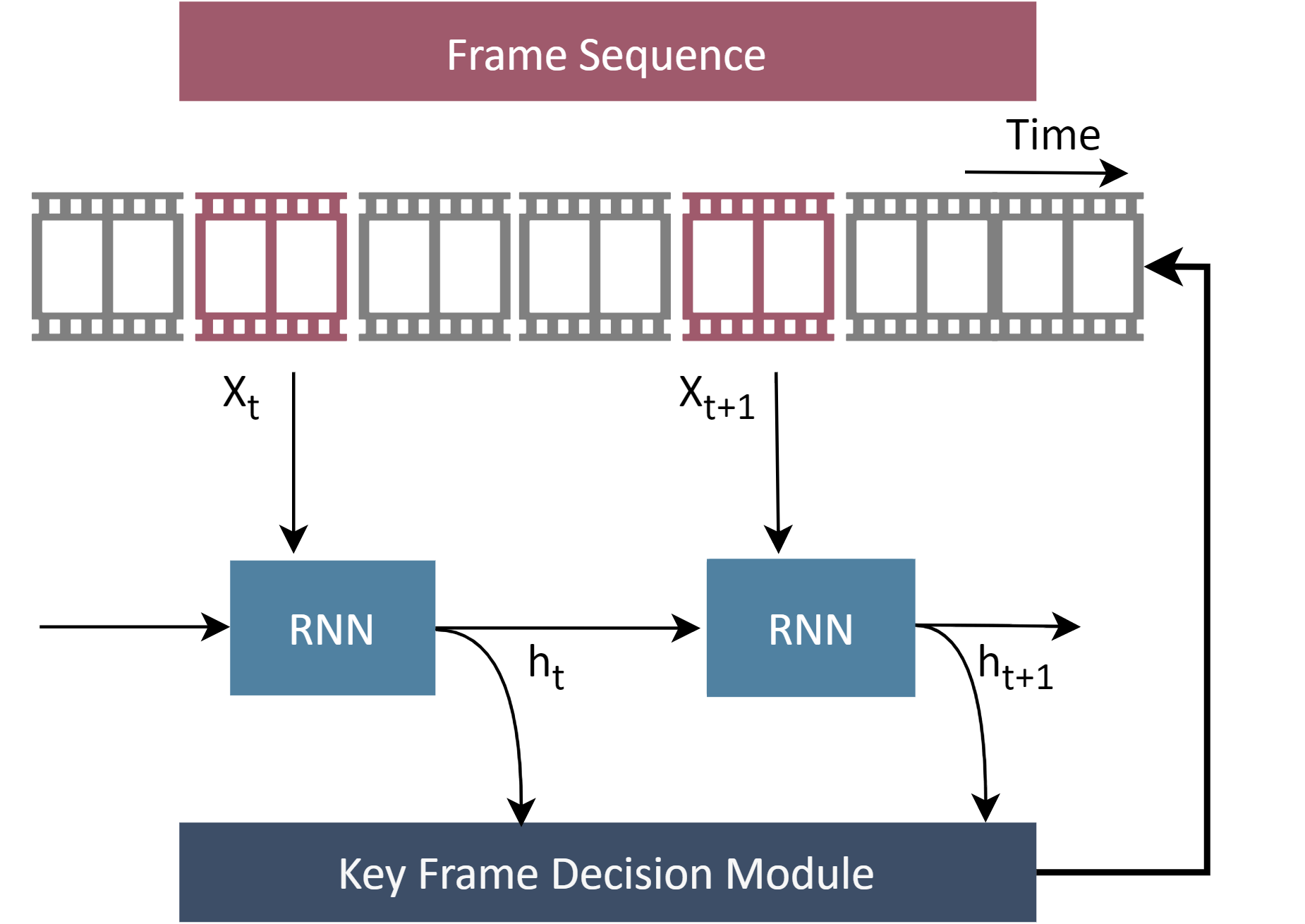}
    \label{frameSample}}
 \end{subfigure}
  \begin{subfigure}[Sequence-wise exiting]{
    \includegraphics[scale=0.057]{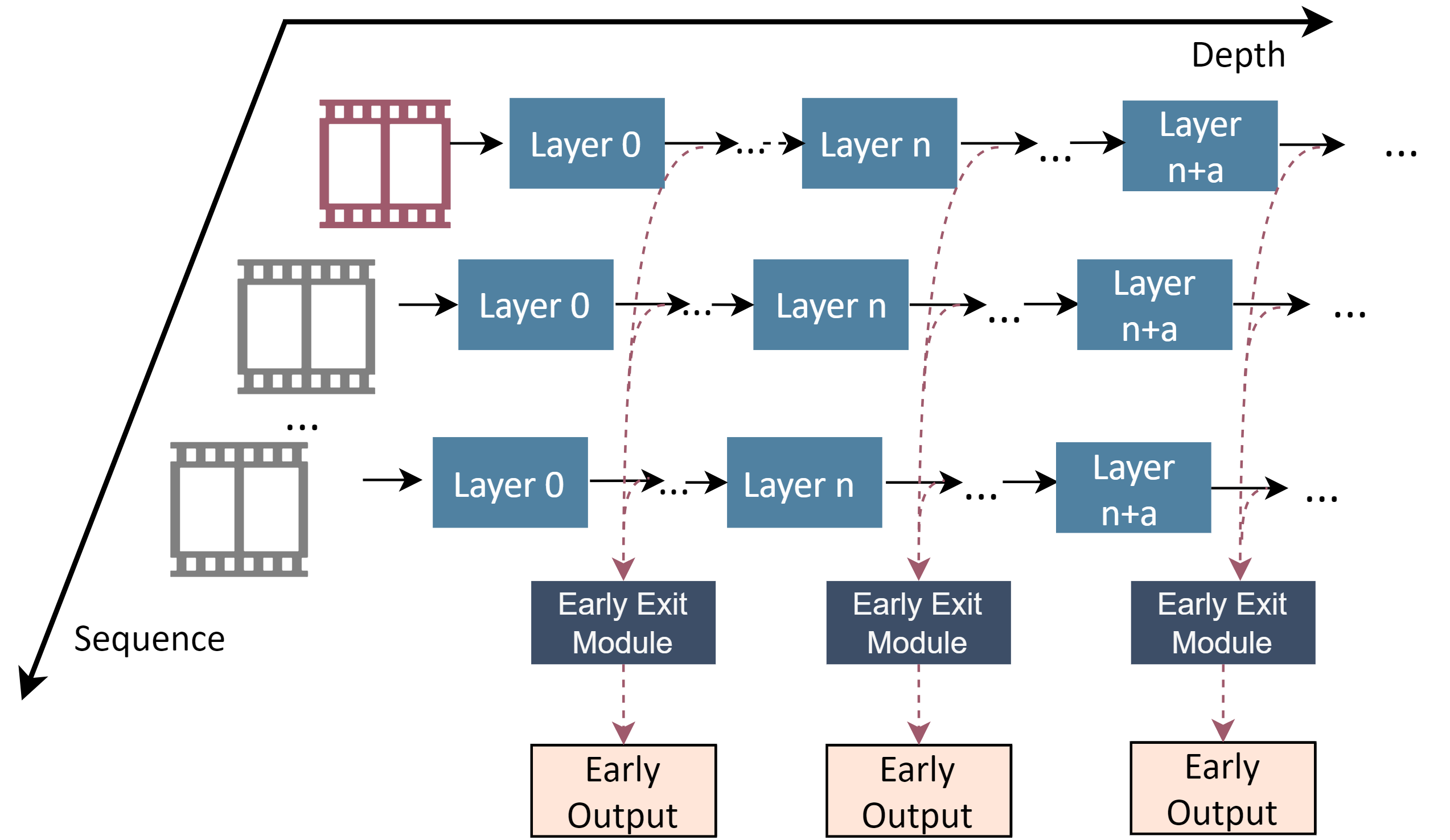}
     \label{earlyexiting}}
  \end{subfigure}
  \caption{DNN-Based temporal dimension optimization focuses on: (a) Feature-wise propagation. Aggregating features from past/future support frames to capture long-range dependencies between frames. (b) Frame-wise sampling. Updating the RNN state $h_t$ with frame input $x_t$ to predict the location and content of key frames. (c) Sequence-wise exiting. Using intermediate predictions in inference as conditions for early exiting.}
  \label{temporal dimension}
 \end{figure}

In certain scenarios, like video understanding, reliance on video objects/scene continuity poses a significant challenge in establishing temporal correlations between video frames. To balance processed quality and speed, there's an urgent need for optimization along the temporal dimension. Recognizing that not all frames contribute equally to the outcome, some works reasonably allocate computing resources on key parts, leveraging temporal details. Next, to address varied granularity needs in capturing temporal details, we summarize these efforts across three different data processing scales: feature-wise (see Fig.\ref{longRange}), frame-wise (see Fig.\ref{frameSample}), and sequence-wise (see Fig.\ref{earlyexiting}).\par

\subsubsection{Feature-wise Propagation}Inter-frame feature propagation involves transmitting and reusing high-dimensional features extracted from key frames by establishing inter-frame long-range dependencies to reduce computational time. Yet, unlike the singular spatial information in a single image, the spatiotemporal characteristics of videos make modeling long-range dependencies of objects/scenes within a frame sequence more challenging. For this, Wang et al.~\cite{18non} introduce non-local neural networks to address the challenge of capturing long-term dependencies by convolutional and recursive operations. The non-local operations weigh the sum of features at all locations in the input feature map in response to a location feature. Guo et al.~\cite{19progressive} discuss the limitations of constructing dependencies using traditional optical flow methods, which introduce additional model complexity and lack the capability to extract high-dimensional features. They propose the temporal consistency-based Progressive Sparse Localized Attention (PSLA) to establish inter-frame dependencies by taking progressively sparse steps within local regions. Video Swin Transformer~\cite{22video} studies spatiotemporal locality, where pixels with closer spatial and temporal distances exhibit stronger correlations. Building on this, it designs a transformer-based model with spatiotemporal locality bias. Meanwhile, some works embed the temporal information into the ROI features. For instance, Shvets et al.~\cite{19leveraging} propose region-level temporal relation blocks that model frame-level dependencies and aggregate rich features from past or future frames, further updating the ROI features. Besides, Gong et al.~\cite{21temporal} recognize the high similarity of regional features within the same object and propose the ROI Align operator. This operator combines the ROI features of the current frame with the most similar ROI features in supporting frames.

\subsubsection{Frame-wise Sampling}Frame sampling refers to selecting a set of key frames from the entire video for processing, aiming to reduce data volume and computational complexity. Korbar et al.~\cite{19scsampler} argue that processing each frame within videos is expensive, as salient frames often interleave with segments of minimal change. They propose SCSample, a lightweight sampling model that takes audio features and compressed video features into account, allowing the analysis model to focus on crucial temporal frames. 
Meanwhile, in some learning-based frame sampling, AdaFrame~\cite{19adaframe} is an adaptive sampling framework that employs LSTM as a proxy and global memory as context preservation. Wu et al.~\cite{19multi} introduce multi-agent reinforcement learning for untrimmed video recognition, adjusting frame selection gradually in parallel Markov decision processes. Additionally, Liu et al.~\cite{19looking} and Yao et al.~\cite{20video} also apply reinforcement learning to the decision of key frames. Diverging from sequential frame sampling, OCSampler~\cite{22ocsampler} is an instance-specific one-shot video condensation strategy. Dynamic-STE~\cite{21efficient} emphasizes fully using sampled frames rather than efficient selection. It adopts a student-teacher model to share sampled frames, using the proposed dynamic knowledge propagation as a dual-model interaction mechanism. Considering that the feature impact of video content changes, Yang et al.~\cite{23feature} approach the adaptive keyframe selection method from the perspective of video frame feature similarity.\par

\subsubsection{Sequence-wise Exiting} Sequence-wise exiting mechanism generates intermediate predictions at video feedforward inference, with high-confidence predictions serving as the termination criterion for achieving early exiting and skipping the remaining computation. Fan et al.~\cite{18watching} construct an adaptive stopping network based on reinforcement learning and recurrent neural networks, which analyze frame features and historical data for confident early exit decisions. Wu et al.~\cite{20dynamic} solve the limited frame coverage in early exit. They construct a two-dimensional grid of input frames and network blocks, utilizing pre-defined checkpoints and prediction modules to select exit points. FrameExit~\cite{21frameexit} accounts for the difference in processing difficulty of the input video and utilizes a self-supervised learning cascade gating module to automatically determine the earliest exit point by adjusting the computational budget based on video difficulty. This approach effectively balances the trade-off between accuracy and computational cost.

\section{Application Layer}\label{sec:Applicaion}
DNN-based video analytics models are maturing and are driving progress across diverse fields. Each field possesses unique requirements for video quality and processing efficiency. This section intends to comprehensively outline the optimization of service efficiency in DNN-based video analysis across various fields, such as transportation, commercial retail, and industrial security, tailored to their specific needs.

\subsection{Transportation Field}\label{AppTrans}
\subsubsection{Surveillance Event Query}\label{AppSEQ}
Surveillance event query conducts statistical analysis on video data across various scenarios, such as identifying empty parking spots or locating vehicles with specific markers in surveillance footage. Given the timeliness queries and the scale of query data, there is an urgent need to improve the efficiency of event queries. To date, many studies have implemented efficient query systems in terms of query types, query size, inference models, and other aspects. For query types, Everest~\cite{21top} supports Top-K queries within statistics of independent video frames. The system leverages the specialized CNN and approximate query processing techniques to accelerate query execution while probabilistically guaranteeing accuracy. For large-scale video queries, DVQShare~\cite{21dvqshare} takes time sharing, space sharing, and logic sharing as core ideas, formulates query plans based on a DNN model and realizes concurrent processing of video queries. VideoStorm~\cite{17live} is an efficient query scheduler that balances query resource quality and delay tolerance. It includes an offline analyzer to profile query resource quality and an online scheduler for resource allocation. The deployment results on Azure clusters demonstrate that it performs better compared to fair resource scheduling. In addition, to accelerate queries through an inference-optimized model, Noscope~\cite{17noscope} cascades dedicated models and difference detectors to model the object behavior and highlight differences between frames faithfully.\par 

Also, some studies have optimized the user retrieval efficiency by standardizing and specifying the query language to lower the threshold while using the video system. For example, BLAZEIT~\cite{18blazeit} creates a declarative query language FRAMEQL, a declarative extension of SQL, proposing two methods for improving aggregation and limited queries. Furthermore, REKALL~\cite{19rekall} generates video annotations from multiple sources and associates them with consecutive spatio-temporal messages. It also provides operators for spatio-temporal tags to synthesize new video events. VIVA~\cite{22optimizing} introduces relational hints, a SQL interface that captures model relations and constraints defined by domain-specific knowledge. \par

\subsubsection{Super-resolution Reconstruction}
Surveillance cameras typically employ low-resolution video formats to conserve cache resources. Yet, intelligent security systems are often equipped with real-time super-resolution reconstruction capability of original videos in response to emergency alerts, which rapidly enhances low-resolution surveillance for tasks such as identity verification or license plate confirmation, thereby strengthening security measures and emergency responses. To improve the efficiency of such reconstruction, existing approaches focus on implicit and explicit temporal modeling. RLSP~\cite{20revisiting} studies the impact of high-frequency details and computational load when acquiring temporal data through motion compensation strategies. It introduces a high-dimensional latent state that implicitly transfers temporal information across frames. RRN\cite{19efficient}, a recurrent residual network, utilizes the previous and current frames as hidden state input and integrates identity mapping within the hidden state to maintain temporal information, achieving a balance between runtime and quality of video super-resolution (VSR). Conversely, ETDM~\cite{22look} investigates how explicit temporal difference modeling influences low-resolution and high-resolution spaces. It efficiently computes temporal differences between adjacent and reference frames, replacing continuous frame input to conserve storage while retaining fine details. Furthermore, in real-life super-resolution applications, DAP~\cite{23fast} addresses the frames alignment and fusion challenge when future frame information is unavailable due to causality and real-time constraints. It introduces a recurrent VSR architecture, leveraging the Deformable Attention Pyramid (DAP) to align information from the recurrent state with the current frame prediction.\par

\subsubsection{Anomaly Detection}
The uneven development of society in recent years has contributed to a rise in public security incidents. To deal with this situation, advancements in video surveillance enable real-time behavior analysis and anomaly detection, significantly preventing crimes and enhancing public safety.\par

In this regard, Dai et al.~\cite{21hybrid} propose a hierarchical multimodal architecture for understanding human behavior in videos by modeling the correlations among continuous depth features. The architecture improves video representation using recurrent neural networks and learns model parameters through a self-critical reinforcement learning mechanism, thereby enhancing the inference between spatio-temporal features. Ngoc et al.~\cite{23efficient} argue that employing advanced video classification techniques for real-time monitoring is impractical due to high computation overhead, so they introduce a lightweight 3D-CNN model that incorporates knowledge extraction and contrastive learning techniques to reduce computation complexity.\par

Furthermore, saliency detection, aimed at identifying the most prominent objects or movements within a video, significantly contributes to recognizing behaviors. Certain techniques leverage information across consecutive frames in a video sequence to establish a spatio-temporal attention model, enhancing the efficient recognition of crucial moments and significant areas. Wang et al.~\cite{17video} devise a deep learning model for efficient video saliency detection, comprising static and dynamic saliency modules. The static model's saliency estimation informs direct spatio-temporal inference by the dynamic model. And then Guo et al.~\cite{19motion} argue that motion cues better signify saliency changes than color cues. They present an optical flow-based saliency detection method combining primary motion vectors with spatial cues like appearance features to localize salient regions. Through multi-cue optimization, their approach improves both temporal and spatial result consistency.\par

\subsubsection{Surveillance Object Tracking}
Object tracking is a crucial component of video analysis applications, with real-time implementation forming a resilient technological foundation for time-sensitive advancements. Especially in autonomous vehicle monitoring systems, real-time object trajectory tracking significantly bolsters safety measures while providing instantaneous reference data for driving decision-making processes. Facilitating real-world trajectory analytics, Liu et al.~\cite{21dynamic}  introduce GIS to map object coordinates from frames to real geographic space. They employ YOLO as the base detector and integrate the DeepSORT algorithm with Kalman filtering for dynamic object tracking. For multiple object tracking (MOT), Liu et al.~\cite{22accelerating} propose an optimization strategy that utilizes a lightweight model across common frames to optimize the temporal dimension while designing a region discriminator to reduce DNN input and optimize the spatial dimension. Moreover, studies show that task interdependencies can accelerate video analysis. Bastani et al.~\cite{20miris} suggest the benefits of combining query processing with object tracking to reduce overall processing time. To reduce the amount of frame processing, they integrate query processing into the tracker and refine trajectories solely for queried objects. \par

\subsubsection{Video Scene Understanding}
Similar to object tracking, real-time scene understanding in video is crucial for time-sensitive AI technologies such as autonomous driving. Tailored for low-power embedded platforms, Tosi et al.\cite{20distilled} propose a lightweight multi-stage prediction network architecture that combines video depth, optical flow, semantics, and motion segmentation to refine semantic understanding. S3-Net~\cite{21s3} considers the temporal correlation of video objects/activities and introduces a fast single-shot segmentation strategy, which quickly segments sub-scenes while extracting temporal semantic features as input to the understanding model.\par 

\subsubsection{Traffic Density Estimation}
Situations like road construction congestion or holiday traffic surges demand timely traffic density estimation and rapid congestion mitigation. However, intricate road networks and real-time traffic fluctuations hinder efficient traffic density estimation. For this, DVCF~\cite{18Urban} adopts DNN as the base model, comprising two primary functional modules: feature extraction and localization model. The framework employs an enhanced Single-Point Multibox Detector (SSD) with a ResNet backbone model for vehicle feature extraction. Additionally, it integrates auxiliary convolutional and pooling layers to determine vehicle locations. This process yields class and location information for each vehicle, prioritizing efficiency and accuracy in density estimation while maintaining robustness.\par

\subsection{Commercial Retail Field}\label{AppCom}
The integration of video streaming and online retail (V2R) has become a dominant shopping paradigm. To meet users' expectations for minimal delays, numerous platforms and technologies are designed to accelerate the retrieval and analysis of product information to enable efficient service. 
For example, addressing the demand for large-scale multimedia handling, Hysia~\cite{20hysia}, a cloud shopping platform, integrates DNNs to match video content that optimize the service efficiency. In addition, to provide viewers with information about in-video objects (e.g., clothes, furniture, food), Vandecasteele et al.~\cite{17spott} speed up the process of video summarization, semantic keyframe clustering, and retrieval of similar objects by DNN-based video analytics technologies to extract product information from related objects within videos.\par
 
\subsection{Industrial Security Field}\label{AppIds}
\subsubsection{Industrial Video Summarization}In industrial monitoring, the constant influx of daily data poses storage challenges when accumulated over extended periods. This accumulation can strain storage capacity and impede crucial video analysis. To manage this large-scale industrial data, video summarization technology reduces storage space by analyzing the temporal redundancy and extracting meaningful segments/frames from the original video. For resource-constrained devices, Muhammad et al.~\cite{19deepres} propose a method involving coarse processing and refinement of video data. Coarse processing techniques filter video streams, minimizing transmitted data while retaining essential information. The refinement technique diverts key frames to a summary generation mechanism, consolidating vital data into a concise summary format. 
And Ajmal et al.~\cite{17human} focus on human trajectory-centered video summarization, utilizing support vector machines and Kalman filters to extract trajectories.\par

\subsubsection{Personal Protective Equipment (PPE) Detection}
Frequent engineering accidents in industrial production caused by inadequate safety measures highlight the need for real-time detection of proper protective equipment usage. Nowadays, studies have optimized the detection efficiency in industrial surveillance from a visual perspective. To overcome the transmission limitations of remote inference, Gallo et al.~\cite{21smart} propose an embedded real-time PPE detection system powered by deep learning. Utilizing NCS2 and YOLOv4/YOLOv4-Tiny, they successfully implement helmet detection, achieving efficient detection outcomes. Additionally, Nath et al.~\cite{20deepPPE} conduct a comparative study involving three YOLOv3-based detection models tailored for various wearable devices.\par

\section{Challenges and Open Issues}\label{sec:Challenges}
To better serve those researchers who are interested in DNN-based video analytics and want to have further study on it, in this section, we discuss research challenges, opportunities and open issues for DNN-based video analytics.\par

\subsection{Emerging AI Hardware Support} 
In recent years, the rapid advancements in hardware technology have facilitated the iterations of existing AI co-processors that support video analysis tasks, such as the TPU v3 to TPU v4~\cite{23tpu}. Based on TPU v3, TPU v4 nearly doubled the number of matrix units (MXUs). Additionally, TPU v4 incorporates a dedicated set of sparse cores and an 8-wide SIMD design to optimize neural network models like Transformers. In ongoing evolution of processor architecture, a key challenge lies in synchronously upgrading the software ecosystem, such as the optimization of development tools and programming frameworks. Further exploration of software upgrade approaches is crucial to explore the computational potential of new processors and drive the development of large-scale cloud-based video analysis. Furthermore, the edge resources and computational capabilities limitations pose challenges in deploying \emph{heavy} DNN-based video analytics models at the edge. In response, some specialized edge computing devices have been introduced, such as NVIDIA Jetson Nano~\cite{nvidia2020toolkit}, Huawei Atlas~\cite{18Huawei}, and Bitmain Sophon Edge Developer Board~\cite{20Sophon}. Their low-powered architecture designs allow for easier attachment to edge environments, providing increased opportunities and advantages for running compute-intensive video analysis models at the edge.

\subsection{Video Foundation Model Technology}
The field of artificial intelligence is undergoing a new paradigm, with foundational models like GPT-4 serving as the basis for various applications. These models can be fine-tuned for specific tasks, enhancing development efficiency and adaptability. However, in the domain of video analysis, research on the video foundation model is still in the early stages~\cite{22internvideo, 22omnivl, 23videomae}. One concerning factor is the time cost of training foundational models (e.g., 2400 epochs for Video MAE on 160k videos~\cite{23videomae}). The time-consuming training process restricts the iteration speed of foundational models and their applicability in future real-time service scenarios. In this situation, an open challenge arises from the imbalance between the training efficiency and practical performance of foundation models. Several approaches can be explored to enhance efficiency through a comparative analysis of the structural disparities between conventional task-specific models and foundational models~\cite{23comprehensive, 23unmasked, 21opportunities}: (1) improving the quality of training data; (2) learning spatiotemporal information through multimodal data; (3) compressing and pruning foundational models. Furthermore, interpretability remains a challenge in foundational models. Developing robust theoretical frameworks to guide the efficiency optimization process is one of the keys focuses for future research.

\subsection{6G-Enabled Edge-Cloud Video Scheduling}
The utilization of edge and cloud resources has been a primary concern in the efficiency optimization of edge-cloud video processing. To address this issue, researchers have been dedicated to designing scheduling algorithms~\cite{16mcdnn, 18deepdecision}, such as static/dynamic scheduling, machine learning-based scheduling, and heuristic scheduling. These algorithms determine the optimal execution location for video tasks (edge or cloud), considering the trade-offs among multiple metrics such as latency, network bandwidth, and energy consumption. Nonetheless, deciding where to allocate video tasks between the edge and the cloud remains challenging under the constraints of multiple metrics. Fortunately, the introduction of Sixth Generation Mobile Communications technology (6G)~\cite{226g} brings new opportunities to edge-to-cloud task scheduling. With its significant frequency bandwidth and high data rates, 6G technology can alleviate network bandwidth limitations in scheduling algorithms, providing more choices and possibilities for optimizing task execution locations.

\subsection{New Application Scenarios} As a form of media, video plays a crucial role in various application scenarios. To meet the growing demands for innovation, there is an open challenge and opportunity to develop scenario-specific optimization technologies for video processing efficiency improvement.\par

\begin{itemize}

\item \emph{Machine Perception}: In robotics applications such as autonomous vehicles and unmanned aerial vehicles (UAVs), real-time understanding of video is vital for safety and reliability, as it enables machines to obtain efficient visual information and comprehensive perception of the environment, facilitating accurate judgments and rapid response~\cite{22robotic}.

\item \emph{Virtual Tourism and Digital Heritage}: Combining with virtual tourism and digital heritage, VR/AR/MR technologies~\cite{22virtual} provide seamless and immersive user experiences via low-latency video interactivity. Ongoing research of DNN-based encoding and decoding techniques for 360-degree video remains a promising direction to achieve high compression efficiency and fast scene transitions~\cite{20survey}.\par

\item \emph{Healthcare}: In healthcare applications, safeguarding patient privacy during medical video analysis on edge or cloud platforms is paramount. However, privacy-preserving techniques such as encryption, anonymization, and de-identification often introduce additional computational overhead, posing a critical trade-off between privacy protection and inference efficiency.\par

\end{itemize}

\section{Conclusion}\label{sec:Conclusion}
Accuracy and efficiency are two important metrics for DNN-based video analytics, for which a number of existing studies have been made. Given that existing surveys focused mainly on the accuracy optimization of DNN-based video analytics, in contrast, this paper summarizes existing studies' work from the perspective of efficiency optimization. More specifically, we make a summary of existing efficiency optimization work for DNN-based video analytics from hardware to application level, with efficient hardware support, system deployment, model design and software implementation included. These aim to reduce computational requirements, improve processing speed, and reduce resource requirements. Finally, we consider and discuss possible problems and challenges for the efficiency optimization of DNN-based video analytics. In summary, we hope this survey can help deepen the understanding of existing work on DNN-based video analytics and provides a useful reference resource for those who want to further investigate the efficiency optimization of DNN-based video analytics.




\end{document}